\documentclass[sigconf]{acmart}
\usepackage{float}
\usepackage{multicol}
\usepackage{multirow}
\usepackage{pifont}
\newcommand{\etal}{\textit{et al.}}
\def\BibTeX{{\rm B\kern-.05em{\sc i\kern-.025em b}\kern-.08emT\kern-.1667em\lower.7ex\hbox{E}\kern-.125emX}}
    
\copyrightyear{2021}
\acmYear{2021}
\setcopyright{acmcopyright}\acmConference[MM '21]{Proceedings of the 29th ACM International Conference on Multimedia}{October 20--24, 2021}{Virtual Event, China} \acmBooktitle{Proceedings of the 29th ACM International Conference on Multimedia (MM '21), October 20--24, 2021, Virtual Event, China}
\acmPrice{15.00}
\acmDOI{10.1145/3474085.3479222}
\acmISBN{978-1-4503-8651-7/21/10}
\begin{document}
\fancyhead{}
%

%

%

%

%
\title{Overview of Tencent Multi-modal Ads Video Understanding Challenge}
%

\author{Zhenzhi Wang}
\affiliation{
    \institution{State Key Laboratory for Novel Software Technology, Nanjing University, China}
}
\email{zhenzhiwang@outlook.com}

\author{Liyu Wu}
\affiliation{
    \institution{Peking University Shenzhen Graduate School, China}
}
\email{lywu0420@hotmail.com}

\author{Zhimin Li}
\affiliation{
    \institution{Huazhong University of Science and Technology, Wuhan, China}
}
\email{lizm@hust.edu.cn}

\author{Jiangfeng Xiong}
\affiliation{
    \institution{Tencent Data Platform,\\ Shenzhen, China}
}
\email{jefxiong@tencent.com}

\author{Qinglin Lu} 
\affiliation{
    \institution{Tencent Data Platform,\\ Shenzhen, China}
}
\email{qinglinlu@tencent.com}
%

\begin{abstract}
Multi-modal Ads Video Understanding Challenge is the first grand challenge aiming to comprehensively understand ads videos. Our challenge includes two tasks: video structuring in the temporal dimension and multi-modal video classification. It asks the participants to accurately predict both the scene boundaries and the multi-label categories of each scene based on a fine-grained and ads-related category hierarchy. Therefore, our task has four distinguishing features from previous ones: ads domain, multi-modal information, temporal segmentation, and multi-label classification. It will advance the foundation of ads video understanding and have a significant impact on many ads applications like video recommendation. This paper presents an overview of our challenge, including the background of ads videos, an elaborate description of task and dataset, evaluation protocol, and our proposed baseline. By ablating the key components of our baseline, we would like to reveal the main challenges of this task and provide useful guidance for future research of this area. In this paper, we give an extended version of our challenge overview. The dataset will be publicly available at https://algo.qq.com/.
\end{abstract}
\ccsdesc[500]{Computing methodologies~Activity recognition and understanding}
\ccsdesc[500]{Computing methodologies~Video segmentation}
\keywords{Multi-modal Video Analysis, Temporal Segmentation, Multi-label Classification.}
\maketitle
\section{Introduction}
\label{sec:intro}
As a leading marketing platform in China, Tencent Advertising Group gathers massive business data, marketing technologies and professional service capabilities. The goal of advertising is to convey the right information to the target customer with an appropriate price. Due to the richer contents, better delivery and more persuasive effects of the video-form ads than the image/text-form ads, the percentage of video-form ads has an explosive growth at the era of 5G. Therefore, the understanding of ads videos becomes significant and urgent. In addition to a simple video classification task, a comprehensive ads video analysis temporally has an even more significant impact on many chains of the ads system. For example, 1) By understanding the detailed content of ads videos, a single ads video can be automatically edited to have various lengths in order to fit different ads platforms; and 2) The fine-grained categories of each video segment rather than the coarse video-level categories can better serve ads recommendation. 

By inspecting the statistics of the characteristics in the massive ads videos from the perspective of presentation forms or narratives, we build a comprehensive taxonomy for categories in ads. For example, some characteristics focus on temporal information such as `golden 3 second'; while some focus on audio information such as `pop music' and `single vocal'; the others focus on semantic information such as `ancient china' and `indoor scenes'. These complex semantic representations motivate us to build a class hierarchy with four dimensions: `presentation form', `style', `place' and `narrative': classes in presentation forms mainly describe the ads videos from the perspective of video making, audio or typesetting, while those in the style from the perspective of age, emotions, theme or character relationships. Other categories in place are about locations and those in narratives are about the temporal structure of ads. 

\begin{figure*}[!t]
   \centering
   \includegraphics[width=18cm, height=4.5cm]{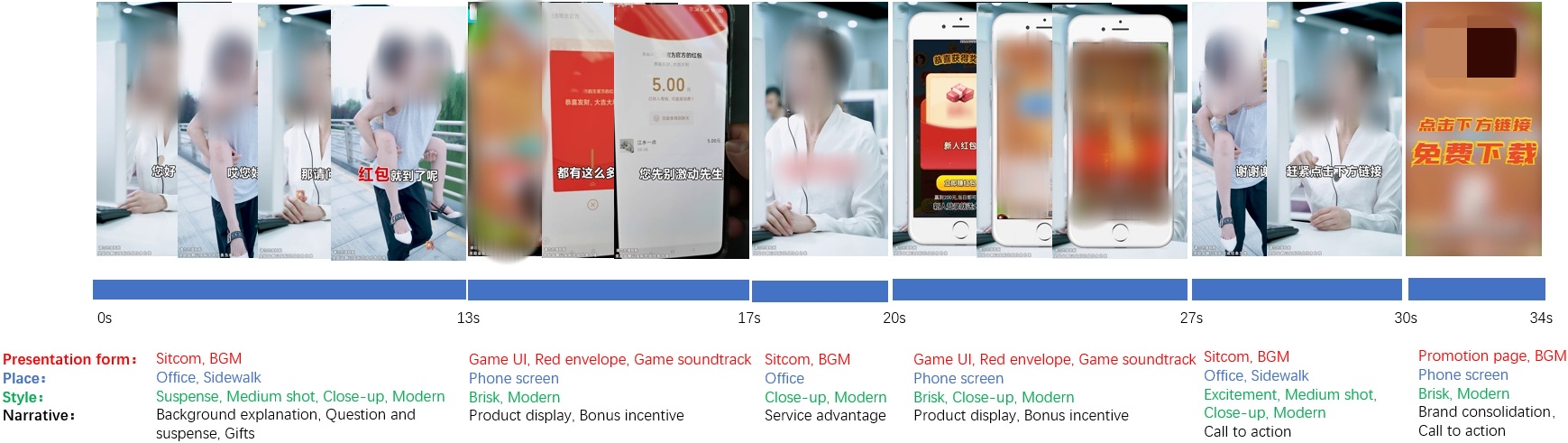}
   \caption{The visualization of our Ads Video Structuring task. By defining semantically-consistent segments as `scene', our task requires precisely segmenting videos as scenes and predicting scene-level multi-label categories.}
\label{fig:first}
\vspace{-1em}
\end{figure*}

\begin{figure}[t]
\centering
\includegraphics[width=0.45\textwidth]{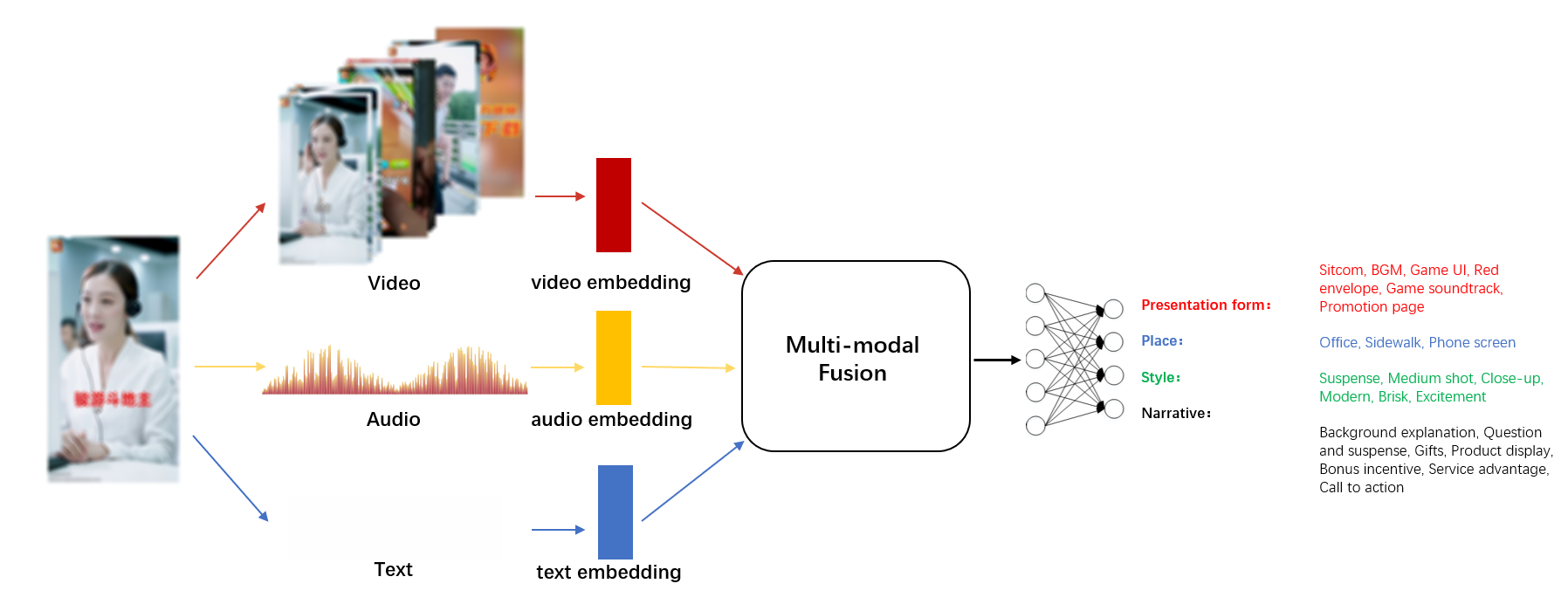}
\caption{The visualization of video classification with multi-modal information. Ads videos commonly contain three modalities: RGB frames, audio, and text (extracted from ASR/OCR). A simple baseline is to extract features from each modality and perform a classification after modality fusion.}
\vspace{-1em}
\end{figure}

Even after the analysis of characteristics and abstraction of ads videos' paradigm, ads video understanding is still too complex for deep models to learn and predict. To make the goal of ads video understanding more specific, we remove the categories which are too abstract and let our challenge focus on three key perspectives: temporal segmentation, multi-modal information, and multi-label classification. Ideally, we want to know the details of each frame in the videos. Yet, the properties of ads videos, such as fast-pace, a large amount of information yet short length, make the gathering of all fragments' details impossible to guide the production of future ads videos. Therefore, we define the concept of `scene', which is a `super shot' in the semantic-level. A scene commonly contains one or some consecutive shots and conveys high-level semantics. For example, the first scene in Fig.~\ref{fig:first} consists of 4 shots, shifting between the couples in the street and the customer service. However, all the 4 shots are about making a phone call, so we group them into a scene. By detecting the boundary of our defined scene, a useful structure of ads video is obtained. Meanwhile, a comprehensive category hierarchy is built, where different categories are related to different modalities. The most important modalities for our task are video frames, audios, and texts obtained from ASR/OCR.

We propose the multi-modal ads video structuring task as the main track of our ACM Multimedia 2021 grand challenge `Multi-modal Ads Video Understanding'. It has three main characteristics: 1) accurate temporal localization of scene boundaries; 2) multi-modal information is essential; 3) a comprehensive category hierarchy from four different perspectives for multi-label scene-level classification. We believe that our task can be a good representative of ads video understanding in the ads system. Based on the analysis of the proposed task above, we collect a large scale ads video dataset with accurate frame-level scene boundary annotations and multi-label category annotations for each scene, named {\em Tencent Ads Video Structuring (Tencent-AVS) Dataset}. This high-quality dataset enables the exploration of our proposed task. Our challenge also has a simplified version as track 2 of our challenge, i.e., video-level multi-label classification task. Next we will mainly introduce track 1 (video structuring task) due to that track 2 is a subset of track 1.

We will give an elaborate description of our proposed benchmark and the baseline model in the following sections: In Sec.~\ref{sec:related}, we will review the topics related to our task and analyze the feasibility of the baseline model. In Sec.~\ref{sec:benchmark}, we will give a detailed description of taxonomy and annotation process of our Tencent-AVS Dataset, and also analyze the proper metrics for our task. In Sec.\ref{sec:baseline} and Sec.\ref{sec:experiment}, we propose our used baseline model and ablate its key components.

\section{Related Works}
\label{sec:related}
Due to the fact that the multi-modal video structuring task is motivated by the need of real-life applications, this task has no yet been explored by the computer vision community before. The most related task is the temporal action segmentation task, yet we have three major differences from it: 1) multi-modal information instead of only RGB frames, 2) there are shot changes in ads videos, and 3) multi-label classification with a long-tailed distribution of categories. In the following subsections, we will first introduce recent progress in temporal video segmentation tasks, and some recent video classification methods as well as utilizing multi-modal information in videos. Then we will analyze the feasibility of several choices for our task's baseline.
\subsection{Temporal Video Segmentation}
Temporally segmenting videos to get the video parts of interest has been explored by the computer vision researchers for several years. A typical setting of video segmentation is temporal action segmentation task which aims to parse the whole video into different human actions. Previous datasets commonly contains instructional videos~\cite{DBLP:conf/cvpr/KuehneAS14,DBLP:conf/huc/SteinM13,DBLP:conf/cvpr/TangDRZZZL019} with human performing diverse actions. Temporal action segmentation requires the model to predict a action label for each frame with only visual information. Recently, another setting of video segmentation is proposed to tackle the movie segmentation problem~\cite{DBLP:conf/cvpr/RaoXXXHZL20}. Different from action segmentation task, movies are usually much longer (e.g., 2 hours ) than instructional videos collected in YouTube (e.g., 5 min), yet temporal segmentation of movies is based on the pre-extracted shots and the evaluation of `scene' segmentation is shot-level instead of frame-level.
\subsubsection{Temporal Action Segmentation}
They usually model the temporal relations and predict action categories for each frame upon densely sampled features extracted by off-the-shelf classification backbones (e.g, I3D~\cite{DBLP:conf/cvpr/CarreiraZ17}). For example, ED-TCN~\cite{edtcn} is a temporal convolution network for action segmentation with an encoder-decoder architecture. TDRN~\cite{tdrn} introduced deformable convolutions and a residual connection into~\cite{edtcn}. MS-TCN~\cite{DBLP:conf/cvpr/FarhaG19} utilized dilated temporal convolution in action segmentation for capturing long-term dependencies and operated it on the full temporal resolution. These works mainly focused on improving receptive field for modeling long-term dependency with encoder-decoder structure~\cite{edtcn,tdrn}, dilated convolution~\cite{DBLP:conf/cvpr/FarhaG19}. Recently, BCN~\cite{DBLP:conf/eccv/WangGWLW20} and ASRF~\cite{DBLP:conf/wacv/IshikawaKAK21} propose to use detected action boundaries to handle over-segmentation errors. Our proposed ads video structuring task contains multi-modal information which cannot be sampled densely, therefore the per frame classification scheme in action segmentation could not apply to ours. Besides, ads videos have a lot of shot change, which may brings more challenges to temporal segmentation methods.
\subsubsection{Scene Segmentation in Movies}
Scene segmentation (or scene boundary detection) with supervised setting is proposed by Baraldi \etal~\cite{baraldi2015deep} and Rotman \etal~\cite{rotman2017optimal} in short videos or documentaries, and scaled up to shot-based long-form movies by Rao \etal~\cite{DBLP:conf/cvpr/RaoXXXHZL20}. In addition, Chen {\it et al.} ~\cite{DBLP:journals/corr/abs-2104-13537} used unsupervised contrastive learning to obtain more discriminative features for scene boundary detection. This task requires temporal models to predict whether each shot boundary is a scene boundary, therefore it is a binary classification problem. This task also involves multi-modal information, such as place, actor, action and audio. Its evaluation is mAP for binary classification without any pre-defined categories. Therefore our task differs from it due to our multi-label classification with a comprehensive category hierarchy from three perspectives.
\subsection{Video Classification}
\subsubsection{Action Recognition with Single-modality}
Video classification is usually investigated in the form of action recognition. Typical methods mainly lie in two categories: 1) two-stream network with RGB frame and optical flow as input to capture appearance and motion information respectively~\cite{simonyan2014two, TSN_ECCV_2016}; 2) 3D convolution networks to directly model appearance and motion with 3D convolution filters, such as I3D~\cite{DBLP:conf/cvpr/CarreiraZ17}, S3D~\cite{xie2018rethinking}, and X3D~\cite{feichtenhofer2020x3d}. Recently, methods with self-attention or transformers~\cite{liu2021video,fan2021multiscale} show advantages in modeling long-range dependency and also achieve state-of-the-art performance on standard benchmarks. However, they mainly focus on human action and model temporal relations via only single modality, i.e., visual features. 
\subsubsection{Multi-modal Information in Videos} 
Recently, multi-modal data is exploited to enrich video representations due to the multi-modal nature of videos. MMT~\cite{gabeur2020multi} uses a multi-modal transformer to jointly encode different modalities in videos, which shows superior performance than previous methods with simple concatenation or channel-wise product as multi-modal fusion. Other methods utilize multi-modal data as supervision signals for pre-training~\cite{sun2019videobert, alayrac2020self, akbari2021vatt}. Different from previous action recognition benchmarks, our proposed ads video structuring dataset has rich multi-modal information such as audio and text extracted from both ASR and OCR. How to effectively utilize multi-modal data in ads videos becomes an challenging problem and deserves further investigations.

\subsection{Feasibility Analysis}
Based on the analysis above, we conclude two possible methods for our ads video structuring task.
\subsubsection{End-to-end Methods.}
End-to-end methods ignore the shot change and directly segment scenes and also classify them with the video features captured by 3D-CNNs~\cite{tran2015learning,DBLP:conf/cvpr/CarreiraZ17} or two-stream network~\cite{TSN_ECCV_2016}. This approach is more practically direct and conceptually simple. Yet it still has some problems: 1) A dense sampling of locations cannot enable the employment of multi-modal information due to the minimum length required by audios or the inaccurate ASR/OCR timestamps, while a sparse sampling may lead to bad performance due to the approximation errors in training and the sparse predictions in inference. The dense sampling of sliding windows will also lead to huge computational costs; 2) The problem of long-tailed distribution becomes even more serious due to the segmented units for classification; 3) Whether the video feature backbone can capture useful informations when crossing the shot change is not sure.

\subsubsection{Multi-stage Methods.}
\label{sec:m-stage}
Due to a lot of shot changes in ads videos, video classification backbones~\cite{tran2015learning,DBLP:conf/cvpr/CarreiraZ17,TSN_ECCV_2016} may be confused when crossing the shot boundaries. To ensure a more stable performance, we decompose the task into three parts and optimize them respectively: 1) shot boundary detection, 2) scene segmentation (a.k.a., scene boundary detection) by aggregating shots, and 3) scene-level classification. 

Firstly, shot boundary detection methods~\cite{DBLP:journals/tcsv/SidiropoulosMKMBT11,DBLP:journals/corr/abs-1906-03363,DBLP:journals/corr/abs-2008-04838} determine whether each frame is a shot change by comparing the difference between adjacent frames. The major difficulty of shot detection is that the gradual transition effects in ads videos are hard to detect. 

Secondly, scene segmentation methods~\cite{DBLP:conf/cvpr/RaoXXXHZL20} commonly use a temporal modeling network such as convolution network or a Bi-LSTM~\cite{schuster1997bidirectional} to capture high-level semantics to aggregate shots as complete semantic units (i.e., `scene' in our task). Different from end-to-end methods, shot-level multi-modal information can be utilized in the scene segmentation stage now due to larger length of the minimum units (i.e., a shot rather than a single frame). 

Finally, a multi-label classification method can be used to predict the categories of each segmented `scene' with multi-modal features. As for multi-modal modeling, a BERT family of models~\cite{DBLP:conf/naacl/DevlinCLT19,DBLP:journals/corr/abs-1910-01108} has been proven as effective to capture semantics of texts; VGGish~\cite{DBLP:conf/icassp/HersheyCEGJMPPS17} uses a CNN architecture to model audio waveforms and achieve promising performance; various video feature backbones mentioned above~\cite{tran2015learning,DBLP:conf/cvpr/CarreiraZ17,TSN_ECCV_2016} can effectively capture both the appearance and motion cues. The features extracted from each modality will be fused in many ways, such as concatenation or cross-attention~\cite{NIPS2017_7181}, to perform the final classification. The major difficulty for scene-level classification is the long-tailed distribution of the categories from four different perspectives. To decouple the influence of different stages and get a more stable performance for our proposed new task, we adopt the multi-stage approach as the baseline of our challenge.

\section{Benchmark}
\label{sec:benchmark}
As mentioned in Sec.~\ref{sec:intro}, we propose the ads video structuring and multi-modal video classification tasks, which we believe is a good representative of ads video understanding. In this section, we will introduce our proposed Tencent Ads Video Structuring (Tencent-AVS) benchmark. To make it solid, we collect large-scale ads videos from real business data and annotate the scene boundaries and the categories with careful quality controls. We also examine many previous metrics and choose proper ones to evaluate methods' performance on our benchmark. We will introduce the details of our dataset in Sec.\ref{sec:dataset} and metrics in Sec.\ref{sec:metrics}.
\begin{figure}[t]
    \centering
    \includegraphics[width=0.5\textwidth]{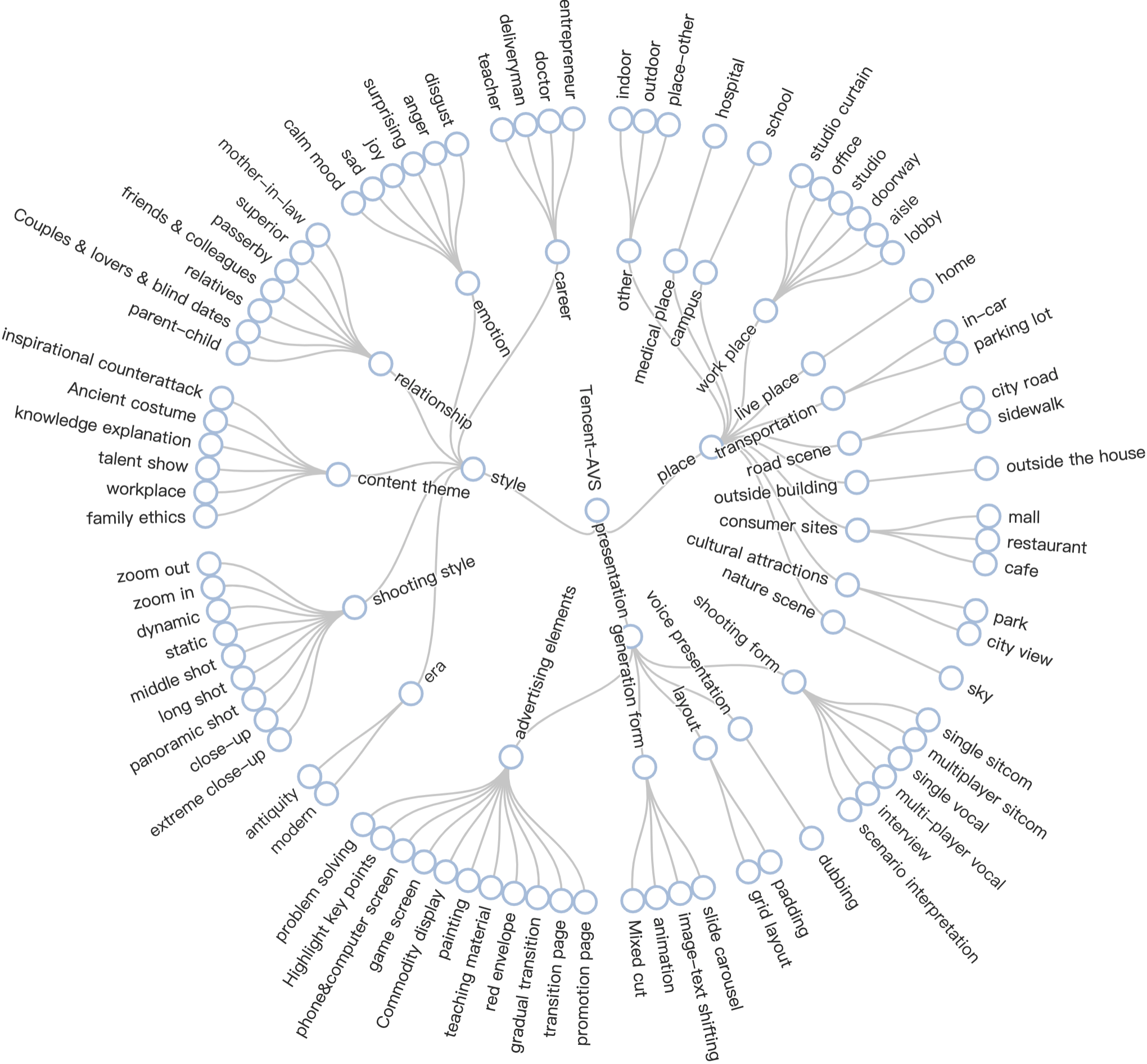}
    \vspace{-1em}
    \caption{The taxonomy tree of Tencent-AVS dataset.}
    \vspace{-1em}
\end{figure}
\subsection{Tencent Ads Video Structuring Dataset}
\label{sec:dataset}
Based on our defined `scene' in Sec.~\ref{sec:intro}, we build our large-scale Tencent-AVS Dataset with accurate annotations for scene boundaries and scene-level multi-label categories. We will introduce the details of our dataset and the annotation process as follows.

\begin{figure*}[t]
    \centering
    \includegraphics[width=1.0\textwidth]{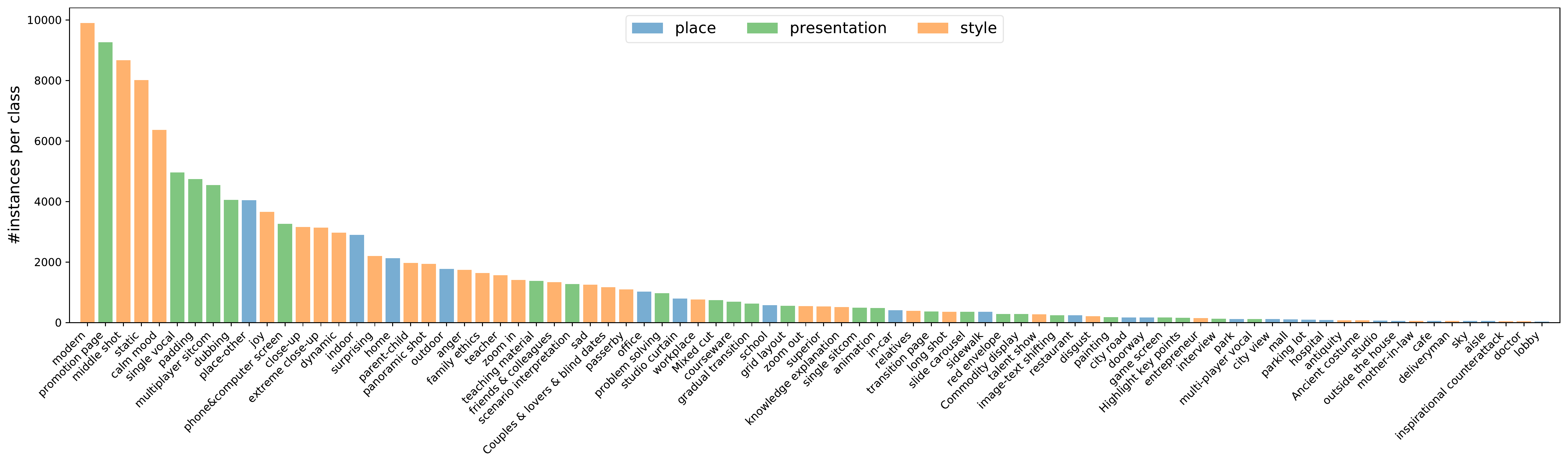}
    \vspace{-2em}
    \caption{Statistics of per class data size. The three dimensions are in different colors (consistent with the hierarchy tree).}
    \label{fig:dist}
    \vspace{-1em}
\end{figure*}

\noindent{\textbf{Taxonomy.}} By exploiting the characteristics of ads videos and collecting the advertiser-uploaded key words, we build a three-layer hierarchy for the categories of ads videos from four perspectives: `presentation form', `style', `place' and `narrative', as shown in Fig.~\ref{sec:dataset}. Then we consult the experts of advertisement and add the missing classes into our category hierarchy to form 233 classes. Finally, we remove the classes with a very small number of instances (i.e., $\leq$ 100) after finishing the annotations of our dataset to make the final dataset have 82 high-frequency classes. The number of category with the presentation, style, and place are 25, 34, and 23, respectively. In the presentation category, there are ad videos covering grid layout, dubbing, and padding. In the style category, ad videos contains emotion, subject, and relationship contents. In the place category, there are locations including home, office, and school. Fig.~\ref{fig:dist} shows the label distribution of presentation, style, and place. We can observe the style contains a large amount of labels and the place contains the least. Besides, the labels among these three main categories suffer form long-tail distributions. An example is showed in the place category where the majority of labels exist within indoor, outdoor, home, and office subcategories.

\begin{table}[t]
\resizebox{0.49\textwidth}{!}{
\begin{tabular}{lllllllll}
\toprule
Split                  & Task                        & \begin{tabular}[c]{@{}l@{}}Total\\ duration\end{tabular} & \begin{tabular}[c]{@{}l@{}}\#Videos\end{tabular} & \begin{tabular}[c]{@{}l@{}}\#Scene\end{tabular} & \begin{tabular}[c]{@{}l@{}}\#Labels\\ per scene\end{tabular} & \begin{tabular}[c]{@{}l@{}}\#Labels\\ per video\end{tabular} & \begin{tabular}[c]{@{}l@{}}\#Labels \end{tabular} \\ \midrule
\multirow{2}{*}{Train} & Track 1    & \multirow{2}{*}{59.2h}                                       & \multirow{2}{*}{5k}                                           & 14.1k                                                  & 5.77                                                                  & 16.29                                                                 & 81.4k                                                 \\
                       & Track 2 &                                                                                     &                                                                  & -                                                      & -                                                                     & 12.14                                                                 & 60.7k                                                 \\ \midrule
\multirow{2}{*}{Val} & Track 1    & \multirow{2}{*}{23.5h}                                       & \multirow{2}{*}{2k}                                           & 5.6k                                                  & 5.77                                                                  & 16.47                                                                 & 32.9k                                                 \\
                       & Track 2 &                                                                                     &                                                                  & -                                                      & -                                                                     & 12.34                                                                 & 24.6k                                                 \\ \midrule
\multirow{2}{*}{Test}  & Track 1     & \multirow{2}{*}{59.4h}                                       & \multirow{2}{*}{5k}                                           & 14.2k                                                  & 5.90                                                                  & 16.80                                                                 & 84.0k                                                 \\
                       & Track 2 &                                                                                     &                                                                  & -                                                      & -                                                                     & 12.49                                                                 & 62.4k                                                 \\  
 \bottomrule                       
\end{tabular}
}
\caption{Statistics of Tencent-AVS dataset.}
\label{tab:stat}
\vspace{-2em}
\end{table}

\begin{figure}[t]
    \centering
    \includegraphics[width=0.49\textwidth]{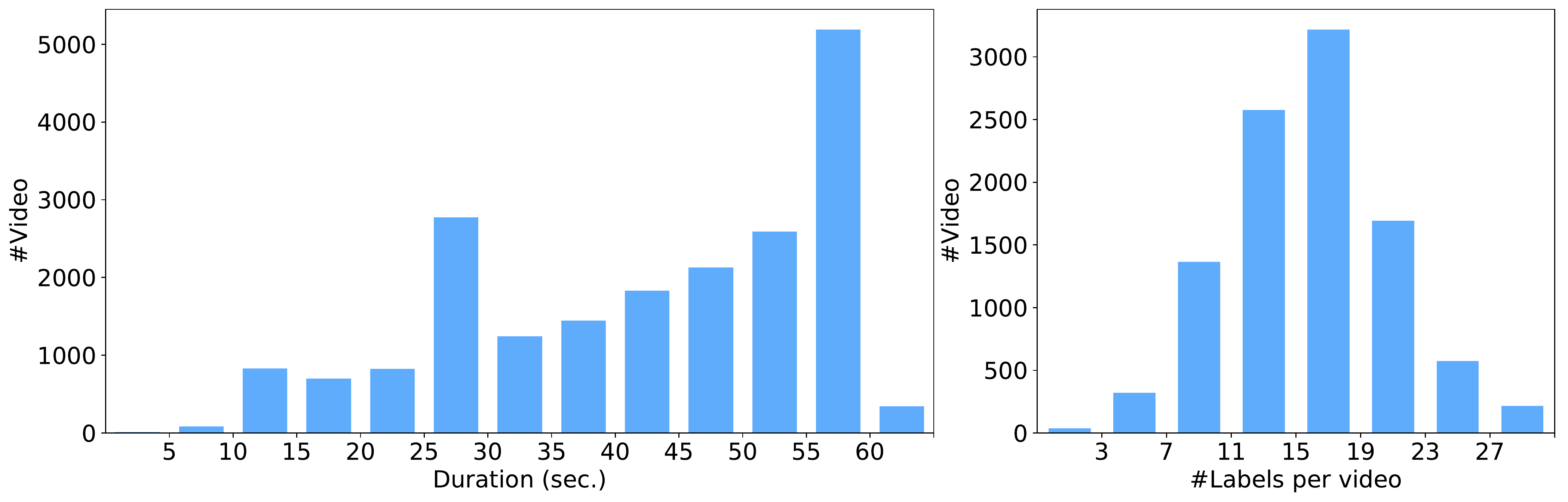}
    \vspace{-1em}
    \caption{(Left) Tencent-AVS video duration distribution. (Right) The average number of labels for each video.}
    \label{fig:duration}
    \vspace{-1em}
\end{figure}

\noindent{\textbf{Annotation process and quality control.}} Dissimilar to the existing action localization or segmentation task whose main challenge of annotation process is the ambiguity of action boundaries, human-edited ads videos have many shot changes as clear demarcations. Therefore, the main challenge of our dataset is the annotations of our multi-label fine-grained categories. Specifically, a scene has about 6 classes on average, so annotators are easy to miss some classes instead of annotating a wrong label for a specific class. Based on these concerns, we write a detailed handbook for annotators, which contains a detailed description and a sample video for each class, and ask them to pay attention to the easily-missed classes. Our handbook also describes when scene changes and in which situation the scene crosses multiple shots. Our annotation process consists of four steps: 1) trial annotation, 2) annotation, 3) sampled inspection and revision, and 4) post-processing. In step 1, we let 7 annotating companies to try the annotation and select top 3 companies to continue step 2 according to their accuracy. After each batch of data is annotated, we will randomly sample about 1/10 videos and check the quality of annotations, especially the missing classes. We will approve the batch if the accuracy of annotation is satisfying (e.g., $\geq$ 90\%) and otherwise let the annotators to revise their annotations for the whole batch. Each video's annotations will be revised for about 3 times on average. Therefore, we believe that the quality of our dataset is carefully controlled. Finally, as scene boundaries annotated by humans are hard to reach the per-frame precision, we use a shot detection method TransNet v2~\cite{DBLP:journals/corr/abs-2008-04838} to refine the little shifting of scene boundaries as a post-processing, i.e., we use the nearest shot boundary to substitute the original scene boundary annotation if the distance between them is small (i.e., $\leq$ 0.1s).

\begin{table*}[t]
\begin{tabular}{r|c|lllllll}
\toprule
\textbf{Dataset} & \textbf{Anno. Type}                   & \textbf{Duration} & \textbf{\#Video} & \textbf{\#Shot} & \textbf{\#Segment} & \textbf{\#Class}  & \textbf{Multi-modal}& \textbf{Domain} \\ \hline
MPII-Cooking~\cite{DBLP:conf/cvpr/RohrbachAAS12}             & \multirow{9}{*}{\begin{tabular}[c]{@{}c@{}}Action\\ segmentation\end{tabular}}         & 9.8h              & 44               & -               & 5.6k               & 65                & \ding{55} & cooking actions                         \\
YouCook~\cite{DBLP:conf/cvpr/DasXDC13}          &                                 & 2.4h              & 88               & -               & -                  & 10                & \ding{55} & cooking actions                         \\
50Salads~\cite{DBLP:conf/huc/SteinM13}         &                                 & 5.4h              & 50               & -               & 996                & 17                & + Depth & cooking actions                         \\
Breakfast~\cite{DBLP:conf/cvpr/KuehneAS14}        &                                 & 77h               & 1.7k             & -               & 8.4k               & 48               & \ding{55} & cooking actions                         \\
"5 tasks"~\cite{DBLP:conf/cvpr/AlayracBASLL16}        &                                 & 5h                & 150              & -               & -                  & 47                & \ding{55}$\dagger$ & instructional videos                        \\
Ikea-FA~\cite{DBLP:conf/dicta/ToyerCHG17}          &                                 & 3.8h             & 101              & -               & 1.9k               & 13                & \ding{55} & assembling furniture                       \\
YouCook2~\cite{DBLP:conf/aaai/ZhouXC18}         &                                 & 176h              & 2k             & -               & 13.8k              &  89              & \ding{55} & cooking actions                         \\
Epic-Kitchens~\cite{damen2018scaling}    &                                 & 55h               & 432              & -               & 39.6k              & 97+300$\ddagger$                & \ding{55}$\dagger$ & cooking actions                         \\
COIN~\cite{DBLP:journals/corr/abs-1903-02874}             &                                 & 476.5h            & 11.8k            & -               & 46.3k              &  180             & \ding{55} & instructional videos                        \\ \midrule
OVSD~\cite{rotman2017optimal}            & \multirow{3}{*}{\begin{tabular}[c]{@{}c@{}}Scene\\ segmentation\end{tabular}} & 10h               & 21               & 10k           & 3k               & -                & + Audio/ASR & Movie                                     \\
BBC~\cite{baraldi2015deep}              &                                 & 9h                & 11               & 4.9k            & 670                & -                & + ASR & Documentary                                       \\
MovieScenes~\cite{DBLP:conf/cvpr/RaoXXXHZL20}      &                                 & 297h               & 150              & 270k            & 21.4k              & -                & + Audio & Movie                                      \\ \midrule
\multirow{2}{*}{\begin{tabular}[c]{@{}c@{}}\textbf{Tencent-AVS}\end{tabular}}      & \multirow{2}{*}{\begin{tabular}[c]{@{}c@{}}{\bf Scene}\\ {\bf sturcturing}\end{tabular}}                           & \multirow{2}{*}{\begin{tabular}[c]{@{}c@{}}\textbf{142.1h}\end{tabular}}            & \multirow{2}{*}{\begin{tabular}[c]{@{}c@{}}\textbf{12k}\end{tabular}}              & \multirow{2}{*}{\begin{tabular}[c]{@{}c@{}}\textbf{121.1k}\end{tabular}}            & \multirow{2}{*}{\begin{tabular}[c]{@{}c@{}}\textbf{33.9k}\end{tabular}}              & \multirow{2}{*}{\begin{tabular}[c]{@{}c@{}}\textbf{82}\end{tabular}}               & \multirow{2}{*}{\begin{tabular}[c]{@{}c@{}}\textbf{+ Audio/ASR/OCR}\end{tabular}} & \multirow{2}{*}{\begin{tabular}[c]{@{}c@{}}\textbf{Advertisement}\end{tabular}}                             \\
&&&&&&&&\\

\bottomrule
\end{tabular}
\caption{Comparison with other video segmentation datasets. Action segmentation datasets annotate action categories for each frame, yet they often have no shot changes and very little multi-modal information. Scene segmentation datasets only annotate scene boundaries but no scene-level categories. Our Tencent-AVS has rich multi-modal information, many shot changes inside scenes, and hierarchical scene-level categories. $\dagger$ means using narrations extracted by ASR during annotation, yet no multi-modal information provided for training/testing. $\ddagger$ Epic-kitchens has 97 verb classes and 300 noun classes.}
\label{tab:compare_track1}
\vspace{-2em}
\end{table*}

\begin{table}[t]
\resizebox{0.49\textwidth}{!}{
\begin{tabular}{r|lllll}
\toprule
\textbf{Dataset}       & \begin{tabular}[c]{@{}l@{}}\textbf{\#Labels}\end{tabular} & \textbf{\#Videos} & \begin{tabular}[c]{@{}l@{}}\textbf{\#Labels}\\ \textbf{per video}\end{tabular} & \textbf{\#Classes} &  \textbf{Domain}                        \\ \midrule
Charades~\cite{sigurdsson2016charades} & 67k             & 9.8k     & 6.8                                                        & 157                    & Actions at Home  \\
ActivityNet~\cite{caba2015activitynet} & 39k         & 28k                       & 1.4          & 203                        & Diverse Actions \\
MPII-Cooking~\cite{DBLP:conf/cvpr/RohrbachAAS12} & 13k                                                          & 273 & 46                                                         & 78               & Cooking          \\
MultiTHUMOS~\cite{yeung2018every} & 38.7k                                                        & 400 & 10.5                                                       & 65                    & Sports           \\ \midrule
\textbf{Tencent-AVS} & \textbf{168.5k}                                                       & \textbf{12k}    & \textbf{14.1}                                                      & \textbf{82}                & \textbf{Ads}        \\ \bottomrule
\end{tabular}
}
\caption{Comparison of Tencent-AVS with other video classification datasets.}
\label{tab:compare_track2}
\vspace{-3.5em}
\end{table}

\noindent{\textbf{Splits and statistics.}} Ads videos in Tencent-AVS dataset come from our real business data. It contains 12k Ads videos (totally 142.1 hours) and is divided into 5,000 for training set, 2,000 for validation set and 5,000 for test set, as shown in Table~\ref{tab:stat}. In Fig.~\ref{fig:duration}, we demonstrate the statistic of the video duration (left) and label instances per video (right) in Tencent-AVS dataset. Most of video duration in Tencent-AVS are between 25 and 60 seconds, with average length as 45.5 sec. As shown in Table~\ref{tab:compare_track1}, Tencent-AVS is comparable to other video segmentation datasets without class annotations (e.g., MovieScenes and Kinetics-GEBD only has scene boundary annotations) and is much larger than video segmentation datasets with class annotations (e.g., 50Salads and Breakfast). Furthermore, our dataset coves a wider range of real-world scenes instead of a very small domain (e.g., cooking), which means our class hierarchy is more diverse than previous datasets. In Table~\ref{tab:compare_track2}, we compare Tencent-AVS with other multi-label video classification datasets: Firstly, The number of instances in our dataset shows a significant advantage than previous datasets. In addition, our dataset contains different types of videos from many industries such as education, games or E-commerce. Finally, our dataset contains not only human actions but also other elements such as places and objects. However, contents from other datasets are from single domain and focus on human actions. 

\noindent{\textbf{Access to our Tencent-AVS dataset.}} Our dataset will be publicly available at https://algo.qq.com: Videos from all three splits and annotations of train/val sets can be downloaded by filling an application sheet. We will maintain an online server for evaluating the performance on the test set and the leaderboard will be updated on the website.
\vspace{-1em}
\begin{figure*}[t]
   \centering
   \includegraphics[width=14cm]{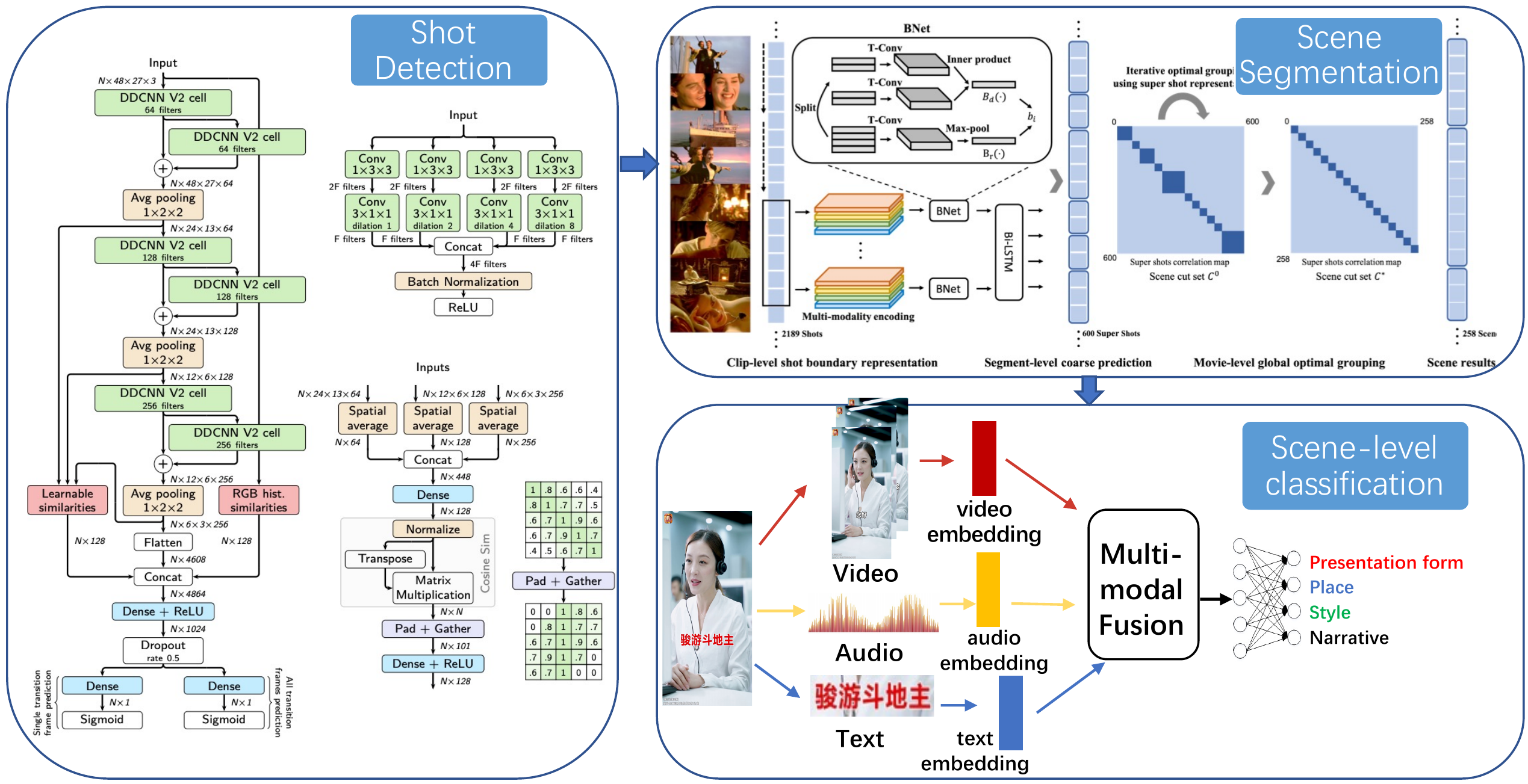}
   \caption{Architecture of our proposed multi-stage baseline.}
\label{fig:baseline}
\vspace{-1em}
\end{figure*}
\subsection{Metrics}
\label{sec:metrics}
\noindent{\textbf{Ads Video Structuring.}} Due to the properties of the proposed task, e.g., multi-label and temporal segmentation, many metrics adopted by existing tasks cannot directly be used in our task. Here we review some related metrics and introduce the metrics we adopted. The most related action segmentation task~\cite{edtcn,DBLP:conf/cvpr/FarhaG19,DBLP:conf/eccv/WangGWLW20} adopts per-frame accuracy, an edit-distance based score and F1-score with tIoU as 0.1, 0.25 and 0.5. However, the multi-label annotation of our task makes these metrics either impossible to compute or unsuitable to use. Some previous scene segmentation methods~\cite{DBLP:conf/cvpr/RaoXXXHZL20} without scene-level class annotations use mAP of scene boundaries as their metric. However, they commonly first use a shot detection method~\cite{shot-detect} and regard the scene segmentation task as a binary classification task which determines whether each shot change frame is a scene boundary. Therefore, it is not suitable for our accurate scene boundary annotations with 0.04s as minimum units. Finally, to measure the performance of our ads video structuring task, we adopt the mAP@tIoU metric similar to ActivityNet challenge~\cite{DBLP:conf/cvpr/HeilbronEGN15} to evaluate the performance of multi-label classification and IoU-based localization in the meantime, and F1@0.5s metric similar to GEBD challenge~\cite{DBLP:journals/corr/abs-2101-10511} to evaluate the accuracy of scene boundary localization. Specifically, for mAP@tIoU, we require the submitted scene segments to have no overlaps for a better adaption for our scene segmentation rather than the original action detection~\cite{DBLP:conf/cvpr/HeilbronEGN15} and take an average of mAPs from IoU 0.5 to 0.95 with stride 0.05 as our final mAP evaluation result; for F1@0.5s, we use 0.5s instead of 5\% of the length of segment in~\cite{DBLP:journals/corr/abs-2101-10511} to avoid the influence from long segments to their neighboring short segments. Formally, the average mAP we use is
\vspace{-1em}
\begin{equation}
m A P[t I o U @ 0.5: 0.95]=\frac{1}{10} \sum_{t I o U=0.5}^{0.95} \frac{1}{N} \sum_{i=1}^{N} A P_{i}^{t I o U}, 
\end{equation}
where $N$ is the total number of categories, and $AP$ is the Average Precision commonly used in detection tasks. Our F1@0.5s matches each predicted boundary to the nearest ground-truth scene boundaries and regard it as True Positive (TP) if their distance is less than 0.5s, otherwise it is a False Positive (FP). The ground-truth scene boundaries will be deleted once it is positively matched (i.e., $\leq$ 0.5s), and the ground-truths which do not have any positive matches to be False Negatives (FN). Therefore, the precision is $TP/(TP+FP)$ and recall is $TP/(TP+FN)$. The final F1@0.5s is computed by 
\begin{equation}
F1@0.5s = \frac{2 \cdot Precision \cdot Recall}{Precision + Recall}.
\end{equation}
\noindent{\textbf{Video-level Multi-label Classification.}} For video-level multi-label classification, we also evaluate the performance by Global Average Precision (GAP)~\cite{kimgoogle}. It calculates the average precision based on top 20 class predictions for each video: 
\begin{equation}
\label{GAP}
    GAP = \mathop{\sum_{j=1}^{n}} p(j)(r(j) - r(j - 1))
\end{equation}
where $n$ is the number of true positive predictions, $p(j)$ is the precision and $r(j)$ is the recall at top $j$.
\section{Baseline}
\label{sec:baseline}

As mentioned in Sec.\ref{sec:m-stage}, we choose the multi-stage method as our baseline, which contains three stages: shot detection, scene segmentation, and scene-level multi-label classification as shown in Fig.~\ref{fig:baseline}. We will introduce the details as follows.

\subsection{Shot Detection} 
We observe that the traditional shot detection method~\cite{shot-detect} shows the bad performance in ads video domain, e.g., having similar colors, gradual transition effects or static images as padding in border regions. Therefore, we use the learning-based method TransNet v2~\cite{DBLP:journals/corr/abs-2008-04838} and optimize its performance for situations such as gradual transition effects or static paddings by adding more training data and a spatial SE~\cite{DBLP:conf/cvpr/HuSS18} module. All the following models are based on the split shots and the scene boundaries are the subset of shot boundaries, so that performance of scene segmentation is bounded by shot detection. Please refer to the TransNet v2 paper~\cite{DBLP:journals/corr/abs-2008-04838} for more implementation details.

\subsection{Scene Segmentation} 
After splitting the videos into a sequence of shots, we adopt LGSS~\cite{DBLP:conf/cvpr/RaoXXXHZL20} to predict the scene boundaries, which is a binary classification problem based on the shot boundaries. Specifically, due to ads videos' shorter duration ($\sim$30s) than the movies in LGSS, we only use the Bi-LSTM~\cite{schuster1997bidirectional} part to model the relations between shots with multi-modal features. The shot-level video and audio features are extracted by pre-trained off-the-shelf feature extractors, i.e., I3D~\cite{DBLP:conf/cvpr/CarreiraZ17} and VGGish~\cite{DBLP:conf/icassp/HersheyCEGJMPPS17}. Then the multi-modal features are concatenated together and fed into LGSS. Please refer to the LGSS paper~\cite{DBLP:conf/cvpr/RaoXXXHZL20} for more implementation details.

\subsection{Multi-model Classification.}
\subsubsection{Feature Embedding Extraction}
We use four different feature extractors to extract video, image, audio, and text embedding, respectively: 1) {\bf Video}: We use a 1 fps sampling rate to process each video. The frames are then sent into ViT-Large~\cite{dosovitskiy2020vit} to be a 1024-d video-level feature vector. 2) {\bf Image}: We take the middle frame of the video as the keyframe, which is processed by ResNet-50~\cite{he2016deep} as a 2048-d feature vector. 3) {\bf Audio}: The audio signal is extracted by FFmpeg and passed through VGGish~\cite{DBLP:conf/icassp/HersheyCEGJMPPS17} network to be a 128-d feature vector. 4) {\bf Text}: Text information is obtained from OCR or ASR. For each sentence, a sequence of word is first tokenized and then fed into BERT \cite{DBLP:conf/naacl/DevlinCLT19} (chinese version). We utilize the 768-d `CLS' token from the last layer of BERT as the sentence feature. The ResNet and Bert networks are finetuned in a end-to-end manner during training while other parameters of feature extractors are fixed.

\subsubsection{Feature Aggregation}
\label{sub:feature-aggregation}
NeXtVLAD~\cite{DBLP:conf/eccv/Lin0018} is efficient and effective method for temporal feature aggregations in videos. By deciding frame-level importance score for a sequence of frames, this network aims to learn the global representations of this sequence. Here we use NeXtVLAD to learn compactly aggregated video representations for visual and audio features.

Specifically, given $M$ frames sampled in a video, we extract the frame-level features $x \in R^{M \times N}$ via a pre-trained backbone. The features are aggregated as $K$ clusters by NeXtVLAD. Each feature is encoded to be a $N \times K$ dimension feature vector $v_{ijk}$ as follows:
\begin{equation}
\vspace{-1em}
\begin{split}
    &v_{ijk} = \alpha_{k}{(x_i)}(x_{ij} - c_{kj})\\
    i\in{\left\{1,\dots,M\right\}},&j\in{\left\{1,\dots,N\right\}},k\in{\left\{1,\dots, K\right\}}
\end{split}
\label{eq:feature1}
\end{equation}    
where $c_k$ is the anchor point of cluster $k$ and $\alpha_{k}(x_{i})$ is an assignment function of $x_i$ to cluster $k$, which measures the similarity of $x_i$ and anchor point of cluster $k$.

In order to save computational cost, NeXtVLAD first expands the feature $x$ as $\dot{x} \in R^{M \times \lambda N}$ via a FC layer, where $\lambda$ is a width multiplier. Then a reshape operation is applied to transform $\dot{x} \in R^{M \times \lambda N}$ to $\Tilde{x} \in R^{M \times G \times \frac{\lambda N}{G}}$, where $G$ is the size of groups. This operation splits $\dot{x}_{i}$ into lower-dimensional features $\{\Tilde{x}_{i}^{g} | g \in \{1, \dots, G\}\}$. Then the feature vector can be written as 
\begin{equation}
\label{eq:feature2}
\begin{split}
    &v^{g}_{ijk} = \alpha_{g}{(\dot{x}_i)}\alpha_{gk}{(\dot{x}_i)}(\Tilde{x}^{g}_{ij} - c_{kj})\\
    g\in{\left\{1,\dots,G\right\}},i\in&{\left\{i,\dots,T\right\}},j\in{\left\{1,\dots,\frac{\lambda{C}}{G}\right\}},k\in{\left\{1,\dots{k}\right\}}
\end{split}
\end{equation}  
where $\alpha_{gk}{(\dot{x}_i)}$ measures the similarity of $\Tilde{x}^{g}_{i}$ to the cluster $k$ and $\alpha_{g}{(\dot{x}_i)}$ can be considered as an attention operation over the groups. $\dot{x}$ and $\Tilde{x}$ are the transformation of input feature $x$, depending on the size of groups $G$, and the width multiplier $\lambda$.  
The video-level descriptor $y$ can be obtained by aggregating frame-level features as
\begin{equation}
\label{eq:feature3}
    y_{jk} = \mathop{\sum_{i,g}}v^{g}_{ijk}
\end{equation}

After we concatenate the global representations of video and audio output by NeXtVLAD, the representation is amplified by a SE Context Gating operation. Finally, a Mixture-of-Experts \cite{ma2018modeling} classifier with a Sigmoid activation is adopted for video-level multi-label classification.

\subsubsection{Loss Function}
\label{sub:loss}
The long-tailed distribution of our dataset brings class imbalance issues. To mitigate this, we follow recent works~\cite{lin2017focal,huang2016imbalance,cui2019class-balanced} to down-weight the loss values for the frequently appeared samples. We utilize asymmetric loss (ASL)~\cite{ben2020asymmetric} ( a improved version of focal loss~\cite{lin2017focal}) to reduce the contribution of negative samples when their predicted confidence are low. Meanwhile, we assign the asymmetric penalty for each positive and negative sample. The losses for positive/negative samples are:
\begin{equation}
    \left\{
\begin{aligned}
L_{+} & = &(1-p)^{\gamma} log(p) \\
L_{-} & = &p_m^{\gamma} log(1-p_m)
\end{aligned}
\right.
\label{eq:asl}
\end{equation}
where $p$ is the network output and $\gamma$ is the penalty parameter for the positive and negative samples. When $\gamma = 0$, Eq.~\ref{eq:asl} degenerates to the binary cross-entropy loss. $p_m=\operatorname{max}(p-m,0)$ is the threshold for discarding easy negatives, which is a lower bound of the confidence $p$. It is worth noticing that the mis-labels, easy negative samples, and dropout make the network converges efficiently. 

\begin{figure*}[t]
    \centering
    \includegraphics[width=1.0\textwidth]{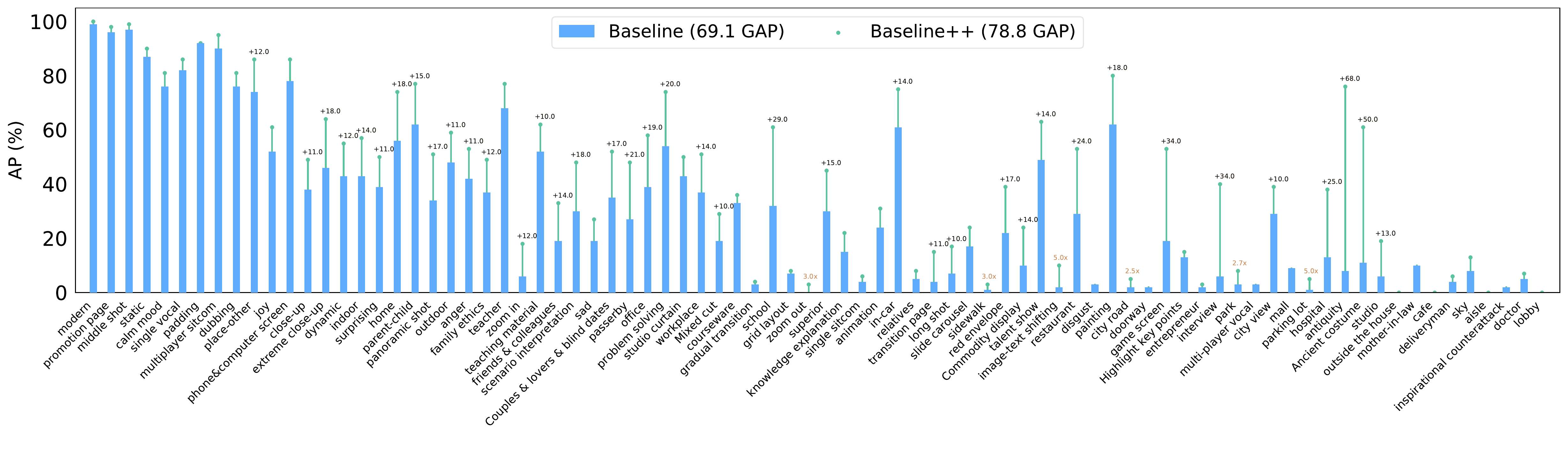}
    \vspace{-2.5em}
    \caption{The comparison of per class AP between the baseline and improved version (baseline++). The order of classes is kept the same with Fig.~\ref{fig:dist} (i.e., descendent order of per class data size).}
    \label{fig:per-class-ap}
    \vspace{-1em}
\end{figure*}

\section{Experiments}
\label{sec:experiment}
To reveal the key challenges of our proposed task, we conduct some important experiments as follows. We hope these experiments could provide useful guidance for the future directions of ads video understanding. All experiments in Sec~\ref{sec:exper_avs} and Sec~\ref{sec:exper_c} follow a standard machine learning pipeline: we use training set to train our model and adjust the hyper-parameters on the validation set, then report the performance on the test set. 
\subsection{Ads Video Structuring}
\label{sec:exper_avs}
\noindent{\textbf{Ablation on temporal modeling in scene segmentation.}} Although Convolution Neural Network has a dominating place in the computer vision community, we observe that LSTM can better perform in the shot-level temporal modeling in Table.~\ref{tab:temporal}. We guess that the reason is the high-level extracted multi-modal features and shorter length sequences of the shot-based modeling rather than commonly-used raw images, so temporal modeling in our task is more similar to NLP tasks. 

\begin{table}[t]
\begin{tabular}{l|cccc}
\toprule
Temporal Model & \multicolumn{1}{c}{Precision} & \multicolumn{1}{l}{Recall} & \multicolumn{1}{l}{F1@0.5s} & \multicolumn{1}{l}{mAP} \\ 
\midrule
CNN            & 78.61                         & 74.06                      &   76.27               &     16.55                     \\ 
Bi-LSTM        & \textbf{82.32}                          & \textbf{74.93}                      & \textbf{78.45}                   & \textbf{20.14}               \\ 
\bottomrule
\end{tabular}
\caption{Ablation on temporal modeling in scene segmentation. The first two metrics are precision and recall of scene boundaries in the shot-level.}
\label{tab:temporal}
\vspace{-2.5em}
\end{table} 
\begin{table}[t]
\begin{tabular}{l|cccc}
\toprule
Modality & \multicolumn{1}{c}{Precision} & \multicolumn{1}{l}{Recall} & \multicolumn{1}{l}{F1@0.5s} & \multicolumn{1}{l}{mAP} \\ 
\midrule
Image (ResNet-50~\cite{he2016deep}) &     63.19                     &        76.96               &     69.40          &               12.23             \\
Audio (VGGish~\cite{DBLP:conf/icassp/HersheyCEGJMPPS17})& 31.04    &    61.55                   &      41.27          & 4.22\\
Video (I3D~\cite{DBLP:conf/cvpr/CarreiraZ17})&80.59&  \textbf{76.25}& 77.83&  19.88\\
All modalities      & \textbf{82.32}                          & 74.93                      & \textbf{78.45}                   & \textbf{20.14}                       \\ 
\bottomrule
\end{tabular}
\caption{Ablation of modalities in scene segmentation.}
\label{tab:seg}
\vspace{-2.5em}
\end{table} 
\begin{table}[t]
\begin{tabular}{l|cccc}
\toprule
Modality &  \multicolumn{1}{l}{mAP} \\ 
\midrule
Image (ResNet-50~\cite{he2016deep})& 14.53\\
Audio (VGGish~\cite{DBLP:conf/icassp/HersheyCEGJMPPS17})& 13.41\\
Video (Inception-v3~\cite{DBLP:conf/cvpr/SzegedyVISW16} + NeXtVLAD~\cite{DBLP:conf/eccv/Lin0018})& 20.00\\
Text (chinese version BERT~\cite{DBLP:conf/naacl/DevlinCLT19})& 2.78\\
All modalities       & \textbf{20.14}                       \\ 
\bottomrule
\end{tabular}
\caption{Ablation of modalities in scene classification. The results are based on the best scene segmentation result.}
\label{tab:cls}
\vspace{-3em}
\end{table} 

\noindent{\textbf{Ablation on modalities in scene segmentation.}} As a multi-modal task, modalities are crucial for recognizing information from different perspectives. We ablate some important modalities in Table.~\ref{tab:seg} and Table.~\ref{tab:cls} to reveal their impact on the final performance: 1) For segmentation, the appearance cues are more important, so the video modality achieves similar results with using all modalities, while merely using audio achieves a very poor result. We think the reason is that audio mainly contains BGM and human voice and it lacks of the discriminative ability to differentiate the scenes. 2) For classification, the text modality contributes the least and audio shows better performance than its in segmentation. Our categories are mainly about appearances or musics yet very few categories are about speech, so the alignment about our categories and text is not good enough.

\noindent{\textbf{Ablation on shot detection methods.}} Since both our temporal modeling and scene classification are based on the shot detection result, its performance has a large impact on our final result. The reason of the large gap between traditional method and deep-learning-based method in Table.\ref{tab:shot} is two-fold: 1) In training, inaccurate shot boundaries will bring noise to the ground-truth of scene boundary in the approximation from the true ground-truth to shot-level binary ground-truth. 2) In inference, models cannot predict a scene boundary if there is no shot boundary.

\begin{table}[t]
\begin{tabular}{l|cccc}
\toprule
Shot detection & \multicolumn{1}{c}{Precision} & \multicolumn{1}{l}{Recall} & \multicolumn{1}{l}{F1@0.5s} & \multicolumn{1}{l}{mAP} \\ 
\midrule
Traditional Method~\cite{shot-detect}     &       50.86                    &        55.23               &          52.95          &     12.21                       \\ 
TransNet v2~\cite{DBLP:journals/corr/abs-2008-04838}$^\dagger$       & \textbf{82.32}                          & \textbf{74.93}                      & \textbf{78.45}                   & \textbf{20.14}               \\ 
\bottomrule
\end{tabular}
\caption{Ablation of shot detection methods.$^\dagger$ means our implementation optimized for ads videos.}
\label{tab:shot}
\vspace{-2em}
\end{table} 

\begin{table}[t]
\begin{tabular}{l|cc}
\toprule
Modality  & GAP \\ \midrule
Video (ViT~\cite{dosovitskiy2020vit} + NeXtVLAD~\cite{DBLP:conf/eccv/Lin0018})           &   76.6  \\
Audio (VGGish~\cite{hershey2017audioset})        &   68.5  \\
Text (chinese version BERT~\cite{DBLP:conf/naacl/DevlinCLT19})           &   58.2  \\
Image (ResNet-50~\cite{he2016deep})       &   63.4  \\ 
All modalities            &   77.1  \\ \bottomrule
\end{tabular}
\caption{Ablation of modalities in ads video classification.}
\label{tab:ab-study}
\vspace{-2em}
\end{table}

\begin{table}[t]
\resizebox{0.47\textwidth}{!}{
\begin{tabular}{l|c|cccc}
\toprule
\textbf{Model} & \textbf{GAP} & \textbf{Backbone} & \textbf{Loss} & \textbf{Classifier} & \textbf{Data Aug.}                                          \\ \midrule
FA$^\ddagger$       & 69.1        & Inception-v3      & CE            & Logistic            & -                                                                   \\ \midrule
     +ViT          & 73.8        & ViT               & CE            & Logistic            & -                                                                   \\
     +ASL          & 76.9        & ViT               & ASL           & Logistic            & -                                                                   \\
     +MoE          & 77.1        & ViT               & ASL           & MoE                 & -                                                                   \\
     +SM$^\ast$         & 75.1        & ViT               & ASL           & MoE                 & SM$^\ast$                                                         \\
     +TM$^\dagger$        & 78.3        & ViT               & ASL           & MoE                 & TM$^\dagger$                                                     \\
      \textbf{+ SM$^\ast$ \& TM$^\dagger$}        & \textbf{78.8}        & \textbf{ViT}               & \textbf{ASL}           & \textbf{MoE}                 & \textbf{SM$^\ast$ \& TM$^\dagger$} \\ \hline
\end{tabular}
}
\caption{Ablation of our baseline (FA). $^\ddagger$FA is feature aggregation. $^\ast$SM is Spatial Mix. $^\dagger$TM is Temporal Mix.}
\label{tab:cls-baseline}
\vspace{-3em}
\end{table}

\subsection{Video-level Classification}
\label{sec:exper_c}
\noindent\textbf{Per-class average precision (AP)}. The per-class AP is shown in Fig.~\ref{fig:per-class-ap}, where all categories are shown in x-axis in a descendent order of per-class data size. We can find that although head classes tend to have better performance, many classes such as `zoom in' and `painting' do not follow this rule, indicating the difficulty of our Tencent-AVS dataset. We denote feature aggregation in~\ref{sub:feature-aggregation} as FA, and FA with asymmetric loss described in~\ref{sub:loss} and data augmentation in~\ref{sub:data} as the proposed method. The blue bar represents the 
performance of FA (i.e., baseline with GAP=$69.1$) and cyan bar is the improved version (i.e., baseline++ with GAP=$78.8$). We achieve a $14.0\%$ relative gain in GAP by our proposed classification method compared to the vanilla FA. Moreover, baseline++ improves the performance of tail classes and maintains the performance of head classes in the same time, indicating the effectiveness of our focus on long-tailed distribution.

\noindent\textbf{Ablation on modalities in video-level classification}. Table~\ref{tab:ab-study} shows the contribution of each modality, where we could observe that the video modality plays the most important role in the multi-modal classification. Although the less important impacts of audio, text and image modality, their performance are complementary. Thus the final result could be further improved by fusion of all modalities.

\noindent\textbf{Ablation on additional components}. Baseline++ contains several modules such as ViT backbone, MoE classifier, ASL loss and data augmentation. Table~\ref{tab:cls-baseline} shows how each module improves the baseline model (i.e., FA). Baseline with inception-v3 achieves GAP $69.1$. By changing the backbone as ViT, the performance is improved to $73.8$. The performance reaches $76.9$ when we add ASL (mentioned in~\ref{sub:loss}). We also use Mixture-of-Experts (MoE)~\cite{ma2018modeling} to improve the classifier, which leads to a $77.1$ GAP. 

\begin{figure}[t]
    \centering
    \includegraphics[width=0.4\textwidth]{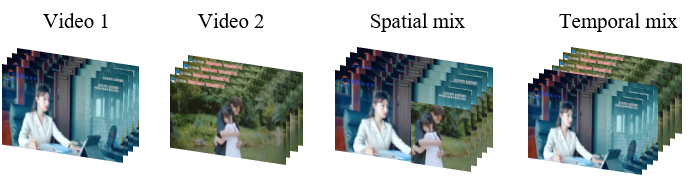}
    \caption{Illustrations of spatial and temporal mix.}
    \label{fig:videomix}
    \vspace{-1em}
\end{figure}

\noindent\textbf{Ablation on data augmentation}. Data augmentation methods are proved to be effective for preventing overfitting and improving generalizations. We follow VideoMix~\cite{yun2020videomix} to conduct two types of data augmentation in videos: spatial mix and temporal mix, as shown in Fig.~\ref{fig:videomix}. We create a new training video which is constructed by replacing and stacking from one video to another. Specifically, spatial mix operation defines a binary mask tensor $M$, which signifies the replace location in two video tensor. Given two video samples $x_1$ and $x_2$, the new video can be obtained by
\begin{equation}
\Tilde{x}=M\odot{x_1} + (1-M)\odot{x_2}
\label{eq:data}
\end{equation}
where $\odot$ is the element-wise multiplication. Similarly, temporal mix operation aims to stack two video along the temporal dimension. 
By further adopting data augmentation methods such as temporal mix (TM) and spatial mix (SM), we achieve a further performance gain on all the additional components and lead to $78.8$ GAP.  

\section{conclusion}
In this paper, we propose the multi-modal ads video structuring task and video classification task in our Multi-modal Ads Video Understanding Challenge to understand ads videos in an in-depth manner. We describe our proposed Tencent-AVS benchmark from the perspectives of class taxonomy, data annotation, and metrics. By ablating our proposed baseline for this task, we reveal the key challenges in ads video structuring and would like to provide useful guidance for future research in this area. We will continue to work with the multimedia community to contribute to related topics.
\bibliographystyle{ACM-Reference-Format}
\bibliography{sample-base}

\end{document}